\pdfoutput=1

\documentclass[11pt]{article}

\usepackage[]{acl}

\usepackage{times}
\usepackage{latexsym}
\usepackage{enumitem}

\usepackage[T1]{fontenc}

\usepackage[utf8]{inputenc}
\usepackage{multirow}
\usepackage{microtype}
\usepackage{xurl}

\usepackage{makecell}
\usepackage{textcomp}
\usepackage{multirow}
\usepackage{caption}
\usepackage{subcaption}
\usepackage{graphicx}
\usepackage{svg}
\usepackage{amsmath}
\usepackage{xcolor}
\PassOptionsToPackage{dvipsnames,svgnames,table}{xcolor}
\usepackage{array}
\usepackage{booktabs}
\definecolor{green}{RGB}{153,255,153}
\definecolor{hred}{RGB}{255,153,153}
\definecolor{darkblue}{RGB}{0,0,153}
\definecolor{teal}{RGB}{0,153,153}
\usepackage{soul}

\newcommand{\PreserveBackslash}[1]{\let\temp=\\#1\let\\=\temp}
\newcolumntype{C}[1]{>{\PreserveBackslash\centering}p{#1}}
\title{Towards Procedural Fairness: Uncovering Biases in How a Toxic Language Classifier Uses Sentiment Information}

\author{ Isar Nejadgholi, Esma Balk{\i}r, Kathleen C. Fraser, and Svetlana Kiritchenko \\
  National Research Council Canada \\
  Ottawa, Canada \\
 \footnotesize \texttt{\{Isar.Nejadgholi, Esma.Balkir, Kathleen.Fraser, Svetlana.Kiritchenko\}@nrc-cnrc.gc.ca}\\
 }

\begin{document}
\maketitle
\begin{abstract}
Previous works on the fairness of toxic language classifiers compare the output of models with different identity terms as input features but do not consider the impact of other important concepts present in the context.  Here, besides identity terms, we take into account high-level latent features learned by the classifier and investigate the interaction between these features and identity terms. For a multi-class toxic language classifier, we leverage a concept-based explanation framework to calculate the sensitivity of the model 
to the concept of \textit{sentiment}, which has been used before as a salient feature for toxic language detection. Our results show that although for some classes the classifier has learned the sentiment information as expected, this information is outweighed by the influence of identity terms as input features. This work is a step towards evaluating procedural fairness, where unfair processes lead to unfair outcomes. The produced knowledge can guide debiasing techniques 
to ensure that important concepts besides identity terms are well-represented in training datasets.

\end{abstract}

\section{Introduction}

Previous NLP works have studied the fairness of toxicity detection classifiers by comparing the distributions of prediction scores across different demographic groups as input features \cite{dixon2018measuring,borkan2019nuanced}. However, other toxicity-related concepts are often present in the text and affect the differences in score distribution between identity groups. Here, we introduce a framework that uses \textit{concept-based global explanations} to uncover unintended biases for different identity groups, while controlling for 
a certain toxicity-related concept. To demonstrate the effectiveness of concept-based explanations in uncovering biases, we specifically focus on \textit{sentiment}, although the general methodology can be applied to any other relevant human-defined concept. Negative sentiment is a salient toxicity feature, which has been used in designing feature-based and neural toxicity detection systems \cite{fortuna2018survey,zhou-etal-2021-hate,chiril2022emotionally}, and highly correlates with toxic language when targeted at demographic groups.

Assessing the differences in score distributions for various demographics is an example of outcome fairness. In fact, most fairness criteria used in machine learning measure outcome fairness, such as accuracy parity (equal accuracy for protected and unprotected groups), equality of opportunity (equal true positive rates), or equalized odds (equal true positive and false positive rates) \citep{morse2021ends}. While valuable, outcome fairness metrics are costly to compute as they require large labelled datasets and do not provide any information about the model's decision making processes. 

More recently, work has begun to focus on the complementary notion of \textit{process fairness} (also known as \textit{procedural fairness}), or the idea that the decision-making process itself must be fair. \citet{grgic2016case} conducted one of the first studies on process fairness in machine learning, measuring the extent to which people believed it was permissible to use various features as input to a criminal recidivism prediction algorithm. For example, they found that people generally felt that \textit{criminal history} was fair to use as an input feature, but that it was unfair to use \textit{family criminality} as input. Another aspect of process fairness is that the importance given to an attribute in the decision-making process shouldn't be very different for different demographic groups. An example of this is the recent \textit{SFFA vs. Harvard} court case where it was argued that academic and extracurricular achievements of Asian-American applicants are given less weight in the admissions process compared to their White-American counterparts \citep{arcidiacono2022asian}. We take a similar view of process fairness and consider a classifier as unfair if it either ignores or over-utilizes a feature for some demographic groups compared to others.

In the current NLP landscape, one major barrier to assessing process fairness is that predictive models rarely use human-understandable concepts as input features, and so it is increasingly difficult to understand what high-level features\footnote{Here, by ``feature'' we mean the latent representations of a semantic concept learned by a classifier, as opposed to the low-level input features.} are actually being learned and used by the classifier. 
In this work, we use an interpretability framework of concept-based explanations \cite{hitzler2022human}, which enables us to explain a machine learning model’s decision-making via conceptual units understandable to humans. 

Concept-based explanations have been studied mostly in the context of computer vision, where it is fairly straightforward to define concepts of interest with a set of representative examples. 
However for textual data, it is much less clear how to define a concept in an effective and intuitive manner, and global explainability methods that operate on high-level abstractions remain under-explored \citep{danilevsky2020survey, balkir2022trustnlp}. \citet{ghorbani2019towards} define a concept to be a meaningful, human-defined abstraction, which is expected to be important for the task at hand and which can be specified by a coherent set of examples. Following this definition, we identify \textit{sentiment} as a concept for toxicity classification.\footnote{
We distinguish between the \textit{concept} of sentiment, as defined by a human through a set of examples, and the \textit{feature} of sentiment, which is implicitly learned by the classifier, although our assumption is that they are closely aligned.} To the best of our knowledge, this is one of the first works to apply concept-based explanations to the domain of NLP, and the first one to explore its effectiveness in identifying high-level fairness issues in models that work with textual data.

In this work, we show how to use concept-based explanations to determine whether a trained toxicity classifier uses the information of \textit{sentiment} as an important feature in its predictions. For that we use a multi-class model, described in Section \ref{sec:model}, and compare the importance of the concept of sentiment in predicting different subtypes of toxicity. Although intuitively, negative sentiment should be an important signal for toxicity detection, its presence is neither necessary nor sufficient for an utterance to be tagged as toxic. 
For example, ``\textit{Muslims are grieving}'' carries a negative sentiment but is not abusive, whereas ``\textit{You are so smart for a woman}'' is perceived as an insult despite including a positive sentiment word. 
Also, sentiment might not be a distinguishing feature for some variations or subtypes of toxic language, such as threats or cyberbullying. For all the classes of our multi-class model, we ask, ``Has the classifier learned the concept of sentiment as a coherent and important high-level feature associated with this label?'', and answer this question with concept-based explanations (Section \ref{sec:sentiment-sensitivity}).
We then assess how the presence of identity terms impacts the use of sentiment information by the classifier. For that, we control the context for sentiment and ask if the learned sentiment information is used similarly and fairly across identity groups (Section \ref{sec:group-sentiment}). Our code and data is available at {\footnotesize  \url{https://github.com/IsarNejad/Procedural-Fairness-Sentiment}}.

\vspace{5pt}
\noindent Our main contributions are:

\setlist{nolistsep}
\begin{itemize}[noitemsep]

\item We propose a concept-based explanation framework to determine whether a trained text classifier uses a human-defined concept fairly in its decision making process. To the best of our knowledge, this is the first work that uses concept-based explanations to uncover biases in text classifiers, and the first to formalize concepts with short textual templates.

  \item To demonstrate the utility of the proposed method, we apply it to a multi-class toxicity classifier and show that when the subject of the sentiment is not specified (e.g.,\@ ``\textit{They are {\small <SENTIMENT-WORD>}''}), the classifier is sensitive to the concept of negative sentiment, for some of the classes.  
  
  \item Further, we show that when the subject of the sentiment is a specific identity term (e.g.,\@ ``\textit{{\small<IDENTITY-TERM>} are {\small<SENTIMENT-WORD>}}''), for some classes, the classifier becomes sensitive to neutral and in some cases even positive sentiment. This demonstrates that the process by which the classifier makes its decision is not the same for all identity groups, and for some groups may even be unfairly associating positive sentiment with toxicity.

\end{itemize}

\section{Multi-Class Toxicity Model }
\label{sec:model}
For our experiments, we use an open-source, RoBERTa-based model\footnote{\url{https://huggingface.co/unitary/unbiased-toxic-roberta}}  \cite{hanu_2020} trained on the English dataset released as part of a Kaggle competition on identifying and reducing bias in toxicity classification of online comments.\footnote{\url{https://www.kaggle.com/c/jigsaw-unintended-bias-in-toxicity-classification/}}. The dataset includes public comments from the Civil Comments platform manually annotated for \textit{Toxicity} as well as six toxicity subtypes: \textit{Severe Toxicity}, \textit{Obscene}, \textit{Identity Attack}, \textit{Insult}, \textit{Threat}, and \textit{Sexual Explicit}. 
The values for each label represent the fraction of the annotators that assigned the label to the comment. There are over 1.8M examples in the training set and around 195K examples in the test set. We exclude the class \textit{Severe Toxicity} from our experiments, since there are only eight training examples with values higher than 0.5 for this class.  
Further, a subset of the data is annotated for various identity groups mentioned in the text. The most frequently mentioned  identity groups include \textit{male}, \textit{female}, \textit{homosexual (gay or lesbian)}, \textit{Christian}, \textit{Jewish}, \textit{Muslim}, \textit{Black}, \textit{white}, \textit{people with psychiatric or mental illness}. 
The classification model optimizes the competition's official evaluation metric that combines the overall AUC with Bias AUCs for the identity groups \cite{hanu_2020}. 
For this, the model's loss function combines the weighted loss functions for two tasks, toxicity prediction and identity prediction. 
This simple and straight-forward model has been shown to effectively reduce bias on non-toxic sentences that mention identity terms, and results in a competitive score of 93.74 on the test set. 

We chose this model for two reasons. First, the model is publicly available and is trained on one of the largest available toxicity dataset, annotated for multiple types of toxicity. An alternative choice for our experiments would be using multiple toxicity classifiers. However, the definitions of subtypes of toxicity are usually ambiguous and similar labels might be used for different subtypes of toxicity across datasets. In the case of our multi-class model, the disparities in using sentiment information can be reliably attributed to differences in sub-type definitions. Second, the model is debiased to some extent with regards to outcome fairness metrics. Uncovering biases in such a model highlights the issue that optimizing for outcome fairness does not guarantee the procedural fairness in decision making.

\section{Sentiment Lexicon}
\label{sec:lexicon}

To formalize sentiment concepts, we employ the NRC Valence, Arousal, and Dominance (NRC-VAD) lexicon \cite{vad-acl2018}, which provides manually annotated real-valued scores of valence, arousal, and dominance for 20,000 English words. 
We use the valence scores and convert them into the range from -1 (the most negative) to 1 (the most positive). 
We automatically select single words from the lexicon that are predominantly used as adjectives in the British National Corpus (BNC)\footnote{The British National Corpus, version 3 (BNC XML Edition),
\url{http://www.natcorp.ox.ac.uk/}} and sort them in decreasing order by their frequency in the BNC. The $N$ most frequent adjectives that can be used to describe humans or groups of humans are manually selected as the sentiment words to define the sentiment concepts.  
The sentiment range [-1, 1] is divided into five intervals: \textit{very negative} [-1, -0.75], \textit{negative} (-0.75, -0.25), \textit{neutral} [-0.25, 0.25], \textit{positive} (0.25, 0.75), and \textit{very positive} [0.75, 1]. 
For each interval, $N = 100$ adjectives are selected.\footnote{The full list of the selected adjectives is available in the Supplemental Material. We also conducted similar experiments with the full NRC-VAD lexicon and obtained similar results.} 
These sets of adjectives are then used to populate the sentence templates to define the sentiment concepts as described in Section~\ref{sec:sentiment-sensitivity}.

\section{Concept-Based Explanations}

Concept-based explanation is an emerging area in black-box model explainability, aiming to explain neural network models at the abstraction level defined by a human user \cite{hitzler2022human}. Most explainability methods provide importance weights for low-level input features such as pixels in images or tokens for text \citep{sundararajan2017axiomatic, smilkov2017smoothgrad, selvaraju2017grad, shrikumar2017learning}. However, a user might want to evaluate the model’s functionality at the level of a concept that is expected to be important for the model’s prediction, which can be achieved with concept-based explanations \cite{koh2020concept}. \citet{ghorbani2019towards} states that a concept needs to satisfy the properties of \textit{meaningfulness}, \textit{coherency} and \textit{importance} for the task at hand. Some examples of concepts in computer vision tasks are the concept of \textit{stripes} for the class of \textit{zebra} \cite{kim2018interpretability}, the concept of \textit{white coat} for the class of \textit{doctor} \cite{pandey_2021}, and the concept of \textit{nuclei texture} in the detection of tumor tissue in breast lymph
node samples \cite{graziani2018regression}. In the case of text, \citet{nejadgholi-etal-2022-improving} used concept-based explanations to measure the sensitivity of a abusive language classifier to the emerging concept of COVID-related anti-Asian hate speech, and \citet{yeh2020completeness} explained a text classifier with respect to the concepts identified through topic modeling.

Here, our goal is to explain the prediction of a toxicity classifier at the level of sentiment information learned by the trained model. Since sentiment is not one of the direct input features of the model, feature importance metrics such as LIME \cite{ribeiro2016should} and SHAP \cite{lundberg2017unified} cannot be used to provide its importance. Instead, we consider each level of sentiment as a concept and calculate the importance of these concepts for the model’s predictions with Testing Concept Activation Vectors (TCAV) algorithm. In the following section, we  explain the TCAV algorithm in detail. 

\subsection{Testing Concept Activation Vectors}


Testing Concept Activation Vectors (TCAV) is an algorithm from the family of concept-based explainability methods, which measures the importance of a human-defined concept for model's predictions \cite{kim2018interpretability}. In TCAV, each concept is defined with a set of examples and represented as Concept Activation Vectors (CAVs), in the activation layer of the trained model. TCAV formalizes the importance of a concept as the fraction of input examples for which the prediction scores of the model increase if the input representation is infinitesimally moved towards the concept representation. The prediction increase is measured by calculating the directional derivatives of the prediction layer to CAVs. To calculate the statistical significance of a concept, multiple subsets of concept examples are used to form multiple CAVs, and a TCAV score is calculated for each CAV. A concept is considered to be important for a class if the distribution of its TCAV scores is significantly different from the TCAV scores of a random concept defined by random examples. 

Here, we explain how the TCAV procedure measures the importance of a concept for a class of a RoBERTa-base classifier, in more detail. Similar to \citet{nejadgholi-etal-2022-improving}, we define each concept $C$ with $N_C$ concept examples, and map them to RoBERTa representations of the [CLS] token ${r_C^j}, j=1,...,N_{C}$. Then, $P$ number of Concept Activation Vectors (CAVs), $\upsilon_C^{p}$, are generated by averaging the RoBERTa representations of $N_{\upsilon}$ randomly chosen concept examples, to represent $C$ in the activation space:

\begin{equation}
  \upsilon_C^{p} = \frac{1}{N_{\upsilon}}\sum_{j = 1}^{{N_\upsilon}}{r_C^j} \quad p= {1,..,P}
\label{eq:CAV}
\end{equation}

\noindent where $N_{\upsilon}<N_C$. With $f_{emb}$, which maps an input text $x$ to its RoBERTa representation $r_x$, the \textit{conceptual sensitivity} of a class to the $\upsilon_C^{p}$, at input $x$ can be computed as the directional derivative $S_{C,p}(x)$:

\vspace{5pt}
$S_{C,p}(x) =
    \lim\limits_{\epsilon \to 0} \frac{h(f_{emb}(x)+\epsilon \upsilon_C^{p}) -h(f_{emb}(x))}{\epsilon}$

\begin{equation}
\quad \quad \quad = \bigtriangledown h(f_{emb}(x)).\upsilon_C^{p} 
\label{eq:sensitivity}
\end{equation}
 
\noindent where $h$ is the function that maps the RoBERTa representation to the logit value of the class of interest. For a set of input examples, $X$, we calculate the TCAV score as the fraction of inputs for which small changes in the direction of $C$ increase the logit:

\vspace{-4mm}
\begin{equation}
    TCAV_{C,p} = \frac{| {x \in X:S_{C,p}(x)>0}|}{|X|}
\label{eq:TCAV}
\end{equation}

When calculated for all CAVs, Equation \ref{eq:TCAV} results in a distribution of scores for the concept $C$. The mean and standard deviation of this distribution determines the overall sensitivity of the classifier to the concept $C$ for the class of interest.

\begin{table*}[!ht]
    \centering
    \small{
    \begin{tabular}{c|c|c|c|c|c|c|c}
    \hline
  \multicolumn{1}{c|}{} & \multicolumn{5}{c|}{Sentiment level concepts }&\multicolumn{2}{c}{Control concepts}\\
\cline{2-8}    
      \multicolumn{1}{c|}{Class label} &Very negative&Negative&Neutral&Positive& Very positive&Explicit&Non-coherent \\
      \hline
      \textit{Toxicity}   &\textbf{0.87 (0.04)}&\textbf{0.47 (0.26)}&0 (0)&0 (0)&0 (0)&\textbf{0.88 (0.02)}&0 (0)\\
      \hline
      \textit{Obscene}  &0 (0)&0 (0)&0 (0)&0 (0)&0 (0)&\textbf{0.75 (0.1)}&0 (0)\\
      \hline
      \textit{Identity Attack}  &0 (0)&0 (0)&0 (0)&0 (0)&0 (0)&0.01 (0.03)&0 (0)\\
     \hline
        \textit{Insult}   &\textbf{0.92 (0.02)}&\textbf{0.77 (0.14)}&0 (0)&0 (0)&0 (0)&\textbf{0.89 (0.02)}&0 (0)\\
        \hline
        \textit{Threat}   &0.01 (0.03)&0 (0)&0 (0)&0 (0)&0 (0)&0 (0)&0 (0)\\
        \hline
        \textit{Sexual Explicit}   &0 (0)&0 (0)&0 (0)&0 (0)&0 (0)&\textbf{0.70 (0.18)}&0 (0)\\
     
     \hline
    \end{tabular}
    \caption{Means and standard deviations of TCAV score distributions for the six types of toxicity with respect to five  sentiment categories and two control concepts. Scores statistically significantly different from random are in bold. }
    \label{tab:sentiment_sensitivity}
    }
\end{table*}

Intuitively, the derivatives in Equation \ref{eq:sensitivity} indicate whether a label's likelihood increases when a small vector in the direction of the concept's representation is added to a random context. For example, the predicted probability of the class \textit{Toxic} for sentence \textit{``I saw these people.''} is $0.01$. The comment \textit{``I saw these people. They are terrible.''} is labeled as toxic with the probability of $0.56$, but the statement \textit{``I saw these people. They are wonderful.''} receives the toxicity probability of $0.01$. If this observation holds systematically across many negative and positive sentiment words, the classifier has learned negative sentiment as an important feature of the toxicity class, but the positive sentiment does not contribute to the toxicity estimation.

In contrast to the previous concept-based explanation works in NLP, which either require annotated data \cite{nejadgholi-etal-2022-improving}, or are limited to the topics extracted by the topic modeling procedure \cite{yeh2020completeness}, we define the sentiment concepts with a set of minimal templates, that are easy to generate and minimize extra contextual information. Using concept examples, described in Sections \ref{sec:sentiment-sensitivity} and \ref{sec:group-sentiment}, TCAV first encodes the information of sentiment in the RoBERTa embedding space. Then, it populates the directional derivatives of the prediction layer with respect to these vectors. If the derivatives are positive for a significant number of the sentiment concept representations, a high average TCAV score is obtained, i.e, sentiment is learned as a coherent and important feature for the label of interest. 

In our implementation, for each concept, 100 CAVs are generated, each of which is the average representation of 50 randomly selected concept examples. For 1000 random input texts (random tweets collected with stop words), the TCAV scores of each CAV are calculated.
Average and standard deviation of TCAV scores are reported to quantify the importance of the concept for this class.

\begin{table*}[!ht]
    \centering
    \small{
    \begin{tabular}{c|c|c|c|c|c|c|c}
    \hline
   \multicolumn{1}{c|}{} & \multicolumn{5}{c|}{Sentiment level concepts }&\multicolumn{2}{c}{Control concepts}\\
\cline{2-8}    
      \multicolumn{1}{c|}{Class label} &Very negative&Negative&Neutral&Positive& Very positive&Explicit&Non-coherent \\
      \hline
      \textit{Toxicity} &\textbf{0.22}&\textbf{0.12}&0.01&0&-0.01&\textbf{0.27}&0.03\\
      \hline
      \textit{Obscene}   & 0.01 &0&0&0&0&\textbf{0.10}&0 \\
      \hline
      \textit{Identity Attack}  &0.01&
0&0&0&0&0.03& 0\\
     \hline
        \textit{Insult}    &\textbf{0.17}&\textbf{0.10}&0.02&0&0&\textbf{0.16}&0\\
        \hline
        \textit{Threat}   &0 &0&0&0&0&0&0\\
        \hline
        \textit{Sexual Explicit}    &0&0&0&0&0&\textbf{0.09}&0\\
     
     \hline
    \end{tabular}
    \caption{Average increase in probabilities when concept templates are added to random texts. Cells in equivalent positions to Table~\ref{tab:sentiment_sensitivity} are in bold. }
    \label{tab:sentiment_prob_increase}
    }
\end{table*}

\vspace{5pt}

\begin{table*}[!ht]
    \centering
    \small{
    \begin{tabular}{c|c|c|c|c}
    \hline
   \multicolumn{1}{c|}{} & \multicolumn{2}{c|}{Increase in Probability }& \multicolumn{2}{c}{TCAV scores}\\
\cline{2-5}    
       \multicolumn{1}{c|}{Class label} &non-coherent&coherent&non-coherent&coherent \\
       \hline
       \textit{Toxicity}&0.11&0.12&0.01 (0.07)&\textbf{0.47 (0.26)}\\
      \hline
        \textit{Insult}&0.09&0.10&0.09 (0.19)&\textbf{0.77 (0.14)}\\
        \hline
    \end{tabular}
    \caption{Average increase in probability and mean and standard deviation of TCAV scores for the non-coherent concept (Very negative or Very positive sentiment) and the coherent concept (Negative sentiment).
 }
    \label{tab:compare-prob-TCAV}
    }
\end{table*}

\section{Classifier Sensitivity to Sentiment}
\label{sec:sentiment-sensitivity}

In this section, we analyze the sensitivity of the classifier described in Section \ref{sec:model} and identify the classes for which the classifier is sensitive to sentiment as a coherent and important feature. For the concepts of sentiment level, we create concept examples and use the TCAV technique to test the importance of the concepts for each class.

To create concept examples for each level of sentiment described in Section \ref{sec:lexicon}, we use the template \@ ``\textit{They are <SENTIMENT-WORD>.}''\@. This simple template ensures minimal extra semantic information other than sentiment level and avoids the problem of encoding unwanted biases in the concept itself, as described by \citet{tong2020investigating}. It is important to note that such minimal templates cannot be labeled for toxicity without more context. Consider a sentence such as \@``\textit{They are terrible.}''. This sentence expresses a negative sentiment but lacks any other significant information. Only with more context can we say whether this sentence is toxic or not. The statement \@ ``\textit{These people are immigrants. They are terrible.}''\@ is toxic, while the comment\@ ``\textit{I don’t like these computers. They are terrible.}''\@ is non-toxic, and ``\textit{I don't like these singers. They are terrible.}'' can be toxic depending on the specific definition of toxicity in a use case. Therefore, the fairness analysis methods that rely on labels cannot be used to study the impact of these templates on the model's predictions.  

In addition to five levels of sentiment, we use two control concepts with predictable sensitivities: 1) A non-coherent concept, defined by a set of random tweets collected with stop words, for which we expect low average TCAV scores for all labels; 2) The concept of \textit{explicit offence} defined by inserting a profane word\footnote{
We use the words from  
\url{https://github.com/chucknorris-io/swear-words/blob/master/en}} in the template  ``\textit{They are {\small<PROFANE-WORD>}}'', for which we expect high sensitivity from at least some of the labels. As the creators of the toxicity model mention, this classifier shows high sensitivity to profanity because of the over-representation of these words in its training data.\footnote{\url{https://github.com/unitaryai/detoxify}}  Table~\ref{tab:sentiment_sensitivity} shows the average and standard deviation of TCAV scores calculated for the seven concepts described above.

We observe that the TCAV scores for the control concepts are as expected---zero sensitivity for a non-coherent, random concept and high sensitivity to the concept of explicit offence for the labels \textit{Toxicity}, \textit{Obscene}, \textit{Insult} and \textit{Sexual Explicit}. For the sentiment concepts, we observe that the classifier is sensitive to \textit{Very Negative} and \textit{Negative} sentiment for the labels \textit{Toxicity} and \textit{Insult}.
\footnote{Intuitively, the classes \textit{Obscene}, \textit{Identity Attack}, \textit{Threat} and \textit{Sexual Explicit} rely on features other than negative sentiment, i.e., profanity, identity terms, violence or intention of harming, and lewdness, respectively.}
We also observe that the classifier is not sensitive to the \textit{Neutral}, \textit{Positive} and \textit{Very Positive} sentiment concepts for any of the labels, which rules out the sensitivity of the classifier to the specific sentence structure of the templates.

 \begin{figure*}
 \centering
          \includegraphics[width=0.8\textwidth, trim={0cm 1.35cm 0cm 0cm},clip]{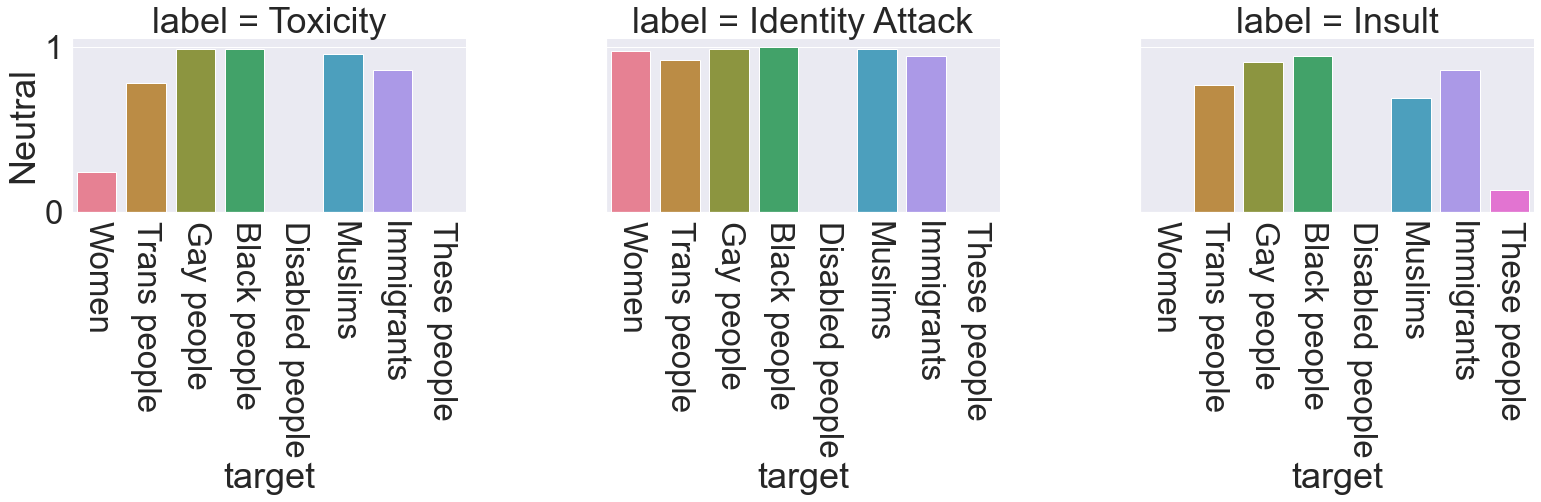}
       \caption{Sensitivity to identity terms in neutral contexts, for which a low sensitivity is expected. }
        \label{fig:subject-sensitivity}
 \end{figure*}

Literature suggests that a high TCAV score indicates: 1) the concept is learned by the classifier as a coherent feature, and 2) that feature is important for the classifiers' predictions \cite{kim2018interpretability}. We evaluate the TCAV scores shown in Table~\ref{tab:sentiment_sensitivity}, in terms of the \textit{importance} and \textit{coherency} of a concept. We first confirm that the \textit{importance} of a concept can be interpreted as the \textit{increase in the predicted probability due to the addition of a concept to input sentences}. Then, we show that \textit{increase in probability} is not an equivalent metric to \textit{TCAV score}, since increase in probability can be due to the addition of a non-coherent concept to input sentences.

\vspace{5pt}

\noindent\textbf{High average TCAV scores indicate a significant increase of prediction probabilities when the concept is added to random contexts}.
We append the concept examples to random tweets and measure the prediction probabilities before and after the addition of the concept examples. The average increase of probabilities for all concepts and labels is shown in Table~\ref{tab:sentiment_prob_increase}. We observe that in all cases where the average TCAV scores are high (i.e., significantly different from the control random concept) in Table~\ref{tab:sentiment_sensitivity}, the probability increase is notable in Table~\ref{tab:sentiment_prob_increase}. 
For example, for the \textit{Toxic} label, the addition of sentences with \textit{Very Negative} and \textit{Negative} sentiment on average increases the prediction probability by $0.22$ and $0.12$, whereas the addition of \textit{Neutral}, \textit{Positive} and \textit{Very Positive} sentiments increases the prediction probability by $0.01$ or less. This is in line with our observation from Table~\ref{tab:sentiment_sensitivity} that the classifier is sensitive to \textit{Negative} and \textit{Very Negative} sentiments for the label \textit{Toxic}.

\noindent\textbf{TCAV scores differentiate between coherent and non-coherent concepts whereas the probability increase does not.} To test this hypothesis, we create a non-coherent concept by combining the \textit{Very Negative} and the \textit{Very Positive} sentiment examples, and compare the average increase in probability and the TCAV score for this concept with those for the \textit{Negative} sentiment concept. The comparison is demonstrated in Table~\ref{tab:compare-prob-TCAV}. Although the increase in probability is similar for the coherent and non-coherent concepts, the TCAV score indicates that the classifier has only learned the coherent concept as an important feature.

\section{Sensitivity to Sentiment Towards an Identity Group}
\label{sec:group-sentiment}

In the previous section, we demonstrated how the TCAV framework can be used to assess whether a human-defined concept is learned by a classifier as an important feature. With that we showed that for some labels our model is sensitive to the presence of \textit{Very Negative} and \textit{Negative} sentiments in broader contexts. Here, we turn to the concept of \textit{``associating a sentiment with an identity group''}\footnote{Note that this concept is composed of more basic concepts, similar to the concept of \textit{white coat} used in \cite{pandey_2021}. Still, it satisfies the three criteria of meaningfulness, coherency and importance as stated by \citep{ghorbani2019towards} and can be considered as a relevant concept for toxicity. } and ask if similar levels of sensitivity to sentiment are observed in the presence of certain demographic terms as input features. For creating the concept examples, we use the template \@ ``{\small<SUBJECTS>} are {\small<SENTIMENT-WORD>}'', where {\small<SUBJECTS>} are the protected identity terms used in HateCheck \citep{rottger-etal-2021-hatecheck}: \textit{Women}, \textit{Gay people}, \textit{Trans people}, \textit{Muslims}, \textit{Immigrants}, \textit{Black people}, and \textit{Disabled people}. We also add the terms \textit{These people} and \textit{These things}, to assess the sensitivity of the model to the concepts of \textit{``associating a sentiment with people in general''}  and \textit{``associating a sentiment with objects''} as two baselines. As expected, we observe that the classifier is not sensitive to any level of sentiment when associated with objects. We discuss some of the most salient results below. (The full results are presented in Appendix in Table~\ref{tab:sentiment_subjects}.)

\begin{figure*}
     \centering
     \begin{subfigure}[b]{0.3\textwidth}
         \centering
         \includegraphics[width=\textwidth,trim={0cm 3.6cm 3cm 0},clip]{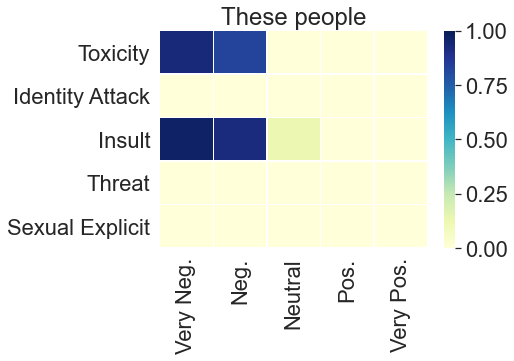}
     \end{subfigure}
     \hfill
     \begin{subfigure}[b]{0.19\textwidth}
         \centering
         \includegraphics[width=\textwidth,trim={5.3cm 3.6cm 3cm 0},clip]{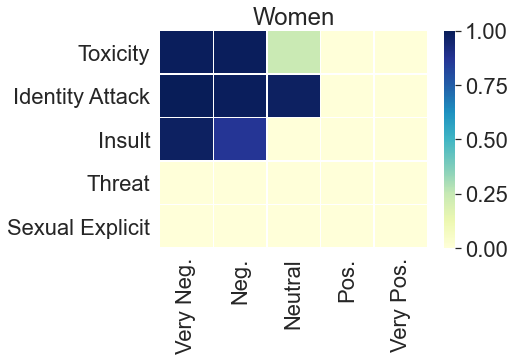} 
     \end{subfigure}
     \hfill
     \begin{subfigure}[b]{0.19\textwidth}
         \centering
         \includegraphics[width=\textwidth, trim={5.3cm 3.6cm 3cm 0},clip]{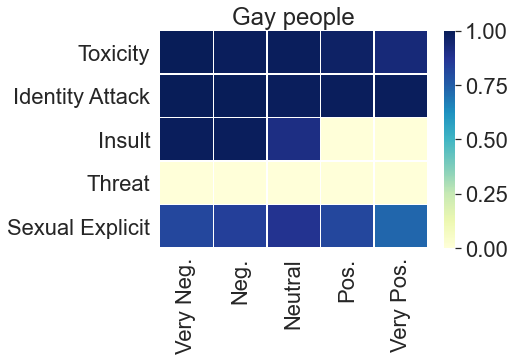}
     \end{subfigure}
     \hfill
     \begin{subfigure}[b]{0.25\textwidth}
         \centering
         \includegraphics[width=\textwidth, trim={5.3cm 3.6cm 0cm 0},clip]{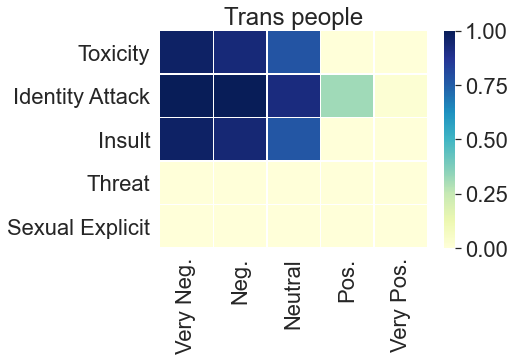}
     \end{subfigure}\\
        
        \begin{subfigure}[b]{0.3\textwidth}
         \centering
         \includegraphics[width=\textwidth,trim={0cm 0cm 3cm 0},clip]{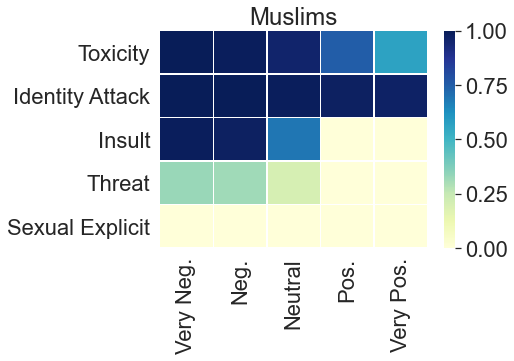}
     \end{subfigure}
     \hfill
     \begin{subfigure}[b]{0.19\textwidth}
         \centering
         \includegraphics[width=\textwidth,trim={5.3cm 0cm 3cm 0},clip]{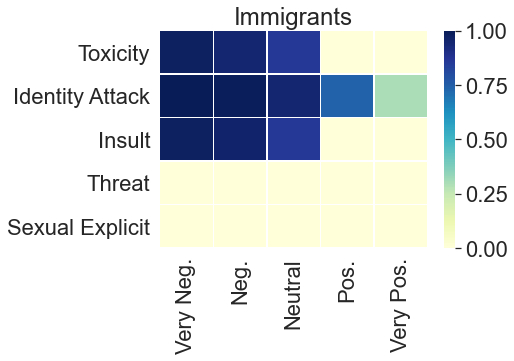} 
     \end{subfigure}
     \hfill
     \begin{subfigure}[b]{0.19\textwidth}
         \centering
         \includegraphics[width=\textwidth, trim={5.3cm 0cm 3cm 0},clip]{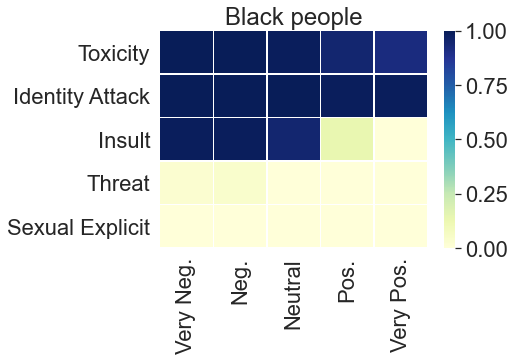}
     \end{subfigure}
     \hfill
     \begin{subfigure}[b]{0.24\textwidth}
         \centering
         \includegraphics[width=\textwidth, trim={5.3cm 0cm 0cm 0},clip]{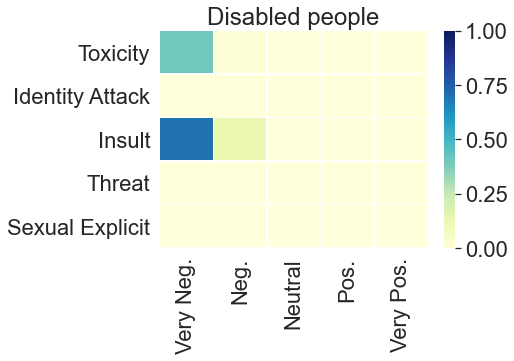} 
     \end{subfigure}
        \caption{Sensitivity to various levels of sentiment for all demographic groups. }
        \label{fig:relative_sentiment}
\end{figure*}

We first assess the influence of identity terms by analysing the classifiers' sensitivities to the neutral sentiment towards the identity groups. Figure~\ref{fig:subject-sensitivity} visualizes the column of Table~\ref{tab:sentiment_subjects} related to the \textit{Neutral} sentiment. From the findings in Table~\ref{tab:sentiment_sensitivity}, as well as human intuition, the association of identity terms with neutral sentiment should not increase the probability of the classifier predicting a toxic label. However, we observe high sensitivities for the labels \textit{Toxicity} and \textit{Insult} and all identity groups, except for \textit{Women} and \textit{Disabled people}. The classifier is also sensitive to the \textit{Neutral} sentiment associated with \textit{Women} for the label \textit{Identity Attack}. We conclude that in neutral contexts the classifier is more sensitive to some identity terms than others.

Figure~\ref{fig:relative_sentiment} visualizes the results of Table~\ref{tab:sentiment_subjects} from a different perspective. In this figure, we assess the sensitivity of the classifier to different levels of sentiment across the identity groups. For the relevant classes, we expect to see that the classifier is sensitive to negative sentiment but not sensitive to positive sentiment; i.e.,\@  the average TCAV score should be high for negative sentiment and low for positive sentiment. Consistent with results in Table~\ref{tab:sentiment_sensitivity}, we observe this expected pattern for the subject \textit{These people} and the two classes \textit{Insult} and \textit{Toxicity}. Taking this as our baseline, we expect to see similar patterns across all the identity groups for an unbiased classifier.
However, we observe that the pattern only holds for certain classes, and certain identity groups. 
Notably, the classifier loses its sensitivity to the \textit{Negative} sentiment for the classes \textit{Insult} and \textit{Toxicity} when the mentioned identity group is \textit{Disabled people}. 
In addition to classes \textit{Insult} and \textit{Toxicity}, in the presence of an identity term, the classifier becomes sensitive to \textit{Negative} and \textit{Very Negative} sentiment for the class \textit{Identity Attack}.  
This is expected given the class definition, but again the pattern does not hold for the identity term \textit{Disabled people}.

Another anomalous result with respect to the \textit{Negative} sentiment is that the classifier is sensitive to \textit{Negative} and \textit{Very Negative} sentiment for the class \textit{Sexual Explicit}, but only in the context of the identity group \textit{Gay people}. Additionally, for the label of \textit{Threat} the classifier shows some sensitivity to the \textit{Very Negative} and \textit{Negative} sentiment, but only when used with the identity term \textit{Muslims}.

We then turn to assessing the sensitivities to positive sentiment. In general, the expected pattern of sensitivities is only observed for the labels \textit{Toxicity} and \textit{Insult} and the identity term \textit{Women}. For other cases, as the level of sentiment changes from \textit{Very Negative} and \textit{Negative} to \textit{Positive} and \textit{Very Positive}, the sensitivity of the classifier does not decrease as expected and remains high, specifically for identity groups \textit{Gay people}, \textit{Black people} and \textit{Muslims}. This observation indicates that a sentence such as ``\textit{Black people are wonderful}'' in a conversation will increase the probability of that text being labeled as \textit{Toxic}, \textit{Identity Attack} and \textit{Insult}. Also, for \textit{Sexual Explicit} the classifier is sensitive to the mention of \textit{Gay people} for all levels of sentiment. We conclude that the classifier is oversensitive to the presence of these identity terms regardless of the level of sentiment, and even highly positive sentiment cannot cancel out the impact of the identity terms. One exception is the term  \textit{Disabled people}, for which the model is under-sensitive, i.e, not sensitive to even \textit{Negative} and \textit{Very Negative} sentiments associated with this group.

\section{Discussion}

Our results demonstrate that while a multi-class toxicity classifier generally shows high sensitivities to negative sentiment for certain classes, and zero sensitivities to neutral or positive sentiment, the picture changes when the sentiment is applied to certain marginalized identity groups. Then, counter-intuitively, even positive sentiment can increase the probability of a toxicity label. This suggests an over-reliance on the identity group term.

Previous work in computer vision has underscored the difficulty in finding unbiased examples with which to define concepts, e.g., when searching for images of men or women to examine gender bias, the examples invariably also contain information about age, race, and so on. Here, in the context of NLP, we propose a generalizable solution to that problem, by generating examples rather than collecting them, and carefully controlling the variable of interest (here, sentiment, although the method could extend to other features). For this we use existing lexicons, without the need to label the examples for the various toxicity classes, as would be required for an analysis of outcome fairness. 

This knowledge of how the model uses the sentiment information can guide debiasing techniques. For example, a data augmentation approach can ensure important features are present in the training dataset. In the case of our model, a data balancing procedure should collect and label positive and very positive sentiments associated with gay people, Black people and Muslims, as well as very negative sentiments associated with disabled people and add them to the training dataset. 

It is important to note that evaluating models for the sensitivity to human-defined concepts is a tool to reveal flaws of a trained model, where prior knowledge about expected sensitivities is available. Similar to previous test suits such as HateCheck \cite{rottger-etal-2021-hatecheck}, our method should not be considered as a standalone evaluation of models. Moreover, observing the expected sensitivities does not guarantee the fairness of the model. Only where unexpected sensitivity patterns are observed, the biases can be detected and mitigated accordingly.

Our method has limitations. We carry our analysis for one grammatical construction that expresses the concept of  \textit{associating sentiment to identity groups}. Future work is needed to assess the generalizability of our results to other expressions of sentiment. Moreover, TCAV requires access to at least some model layers and cannot be applied when the model itself is unavailable.

\section{Related work}

Identifying and mitigating unintended biases in NLP systems to ensure fair treatment of various demographic groups has been focus of intensive research in the past decade \cite{blodgett2020language,shah2020predictive}. 
Various metrics to quantify biases in system outputs have been proposed, including group fairness metrics and individual fairness metrics \citep{castelnovo2021zoo,czarnowska2021quantifying}. 
However, to apply such metrics, the datasets need to be annotated with demographic attributes, which is costly and sometimes infeasible to do (e.g., the demographics of the authors of social media posts are often unknown). 
Alternatively, the bias metrics are applied on synthetic data automatically generated using simple templates \cite{kiritchenko2018examining,borkan2019nuanced}. 
In both cases, the test data are limited, and the evaluation is restricted to a set of pre-defined contexts. 

Explainability techniques (XAI) can potentially help in discovering and quantifying biases. Much work on XAI has been motivated by the need to assist in bias detection and mitigation \cite{doshi2017towards,das2020opportunities}. However, only a handful of NLP studies have actually employed explainability methods for bias detection and to a limited extent  \cite{prabhakaran2019perturbation,kennedy2020contextualizing,aksenov2021fine,balkir-etal-2022-necessity}. \citet{balkir2022trustnlp} surveyed the works at the intersection of fairness and XAI in NLP and discussed conceptual and practical challenges in applying current explainability approaches for debiasing NLP models. Multiple outlined issues stem from the fact that most current XAI methods employed in NLP provide explanations on a local level through post-hoc processing, and it is still an open question how to generalize these local explanations to reveal systematic model biases. The TCAV framework used in this paper produces global explanations and can therefore uncover unfair processes in the model's decision making.

Probing classifiers are well-known interpretability tools used to examine the encoded information in the representation layers of NLP models \cite{conneau2018you}. Probes are trained independently from the original model to predict an externally defined property (e.g., linguistic properties such as part of speech) from the model’s representations. Despite being widely used, several studies revealed that probes are not well-controlled, and caution should be taken when drawing behavioural conclusions about the original model from the performance of probing classifiers \cite{belinkov2022probing}. Also, probes can only assess whether the information about the property of interest is encoded in the representations (e.g., \cite{tenney2019you,rogers2020primer}) but do not provide evidence about how the model uses this information. To that end, extensions of probing classifiers were proposed, which assess  the effect of removing the property's information with counterfactual interventions to provide causal explanations and mitigate biases in NLP classifiers \cite{ravfogel2020null, elazar2021amnesic}. However, several concerns are raised about the effectiveness and the unintended consequences of removing attributes \cite{kumar2022probing}. While causality-driven probing methods assess the necessity of the property for the classifier’s decision, TCAV determines whether the model uses the encoded information as an important signal for a particular class. Also, TCAV allows us to quantify the relative importance of different properties encoded in the representation, which is not feasible with probing classifiers.

The TCAV framework has been developed and mostly applied in image classification. 
In the original paper, \citet{kim2018interpretability} showed how gender and racial biases can be discovered with TCAV in image classifiers. 
\citet{wei2021analysing} extended the method to regression problems, and applied it to detect gender and first language biases in automatic spoken language assessment. 
\citet{tong2020investigating} studied the effectiveness of TCAV in discovering gender biases in image classification and discussed the difficulties in obtaining quality examples to represent a concept while not introducing new sources of bias (e.g., introducing a racial bias when selecting gendered examples).   
\citet{adhikari2021fair} used TCAV to measure gender bias when classifying faces as young or old, and discussed the difficulty of defining `disentangled' concepts that only encode the concept of interest. 
To the best of our knowledge, our work is the first in applying TCAV to discover biases in text classifiers.

\section{Conclusion}

Building on previous studies that measured group fairness in toxic language detection, this work is a step toward a more systematic and fine-grained analysis of procedural fairness in neural model's predictions. We use a global explainability metric to uncover the disparities in how the classifier learns to associate identity terms with domain-relevant concepts, e.g. sentiment. 
Future work will focus on extending the analysis to other concepts known to be important to toxic language detection (profanity, threats of violence, dehumanizing or othering language, and so on) as well as additional classifiers, domains, and types of bias.  

\section*{Ethical Statement}

The presented framework aims to identify fairness issues in text classifiers when identity terms are mentioned in the text. 
As stated above, such evaluation cannot attest for the absence of any biases, but can indicate potential areas of concern. This framework is a complementary approach to other methods of bias detection that are based on the notion of outcome fairness (e.g., using fairness metrics on held-out test sets annotated for mentions of demographics or on specifically designed test suits, such as HateCheck). 
The proposed method cannot be applied to assessing fairness on texts \textit{written by} different demographic groups.

The method requires the identity groups of interest to be specified in advance. 
In the current study, we have included several protected groups, but the list is by no means exhaustive. More protected groups should be included in the future. 
Additionally, it is known that the label used to refer to a social group can itself communicate bias (consider, for example, the difference between \textit{immigrants} versus \textit{migrants} versus \textit{expats}) \citep{beukeboom2019stereotypes}. We have not analyzed the effect of this form of bias on the explanations here.
Furthermore, other, legally non-protected groups (e.g., based on physical appearances, education, etc.), should also be considered as we strive towards inclusive and safe online spaces.

As most AI technology, this approach can be used adversely to exploit the system's vulnerabilities and produce toxic texts that would be undetectable by the studied classifier. 
Specifically, for methods that require access to the model's inner layers, care should be taken so that only trusted parties could gain such access. The obtained knowledge should only be used for model transparency purposes, and the security concerns should be adequately addressed. 

Regarding environmental concerns, contemporary NLP systems based on pre-trained large language models, such as RoBERTa, require significant computational resources to train and even fine-tune. Larger training datasets, such as the one used in this study with almost 2M training examples, used for fine-tuning, usually result in a better classification performance, but also an even higher computational cost. 
To lower the cost of this study and its negative impact on the environment, we chose to use an existing, publicly available classification model.

\bibliography{anthology,custom}

\begin{thebibliography}{49}
\expandafter\ifx\csname natexlab\endcsname\relax\def\natexlab#1{#1}\fi

\bibitem[{Adhikari(2021)}]{adhikari2021fair}
Rittika Adhikari. 2021.
\newblock Fair-doctor: Detecting and mitigating unfairness in neural networks.
\newblock Master's thesis, University of Illinois Urbana-Champaign.

\bibitem[{Aksenov et~al.(2021)Aksenov, Bourgonje, Zaczynska, Ostendorff,
  Schneider, and Rehm}]{aksenov2021fine}
Dmitrii Aksenov, Peter Bourgonje, Karolina Zaczynska, Malte Ostendorff,
  Julian~Moreno Schneider, and Georg Rehm. 2021.
\newblock Fine-grained classification of political bias in {G}erman news: A
  data set and initial experiments.
\newblock In \emph{Proceedings of the 5th Workshop on Online Abuse and Harms
  (WOAH 2021)}, pages 121--131.

\bibitem[{Arcidiacono et~al.(2022)Arcidiacono, Kinsler, and
  Ransom}]{arcidiacono2022asian}
Peter Arcidiacono, Josh Kinsler, and Tyler Ransom. 2022.
\newblock Asian american discrimination in harvard admissions.
\newblock \emph{European Economic Review}, 144:104079.

\bibitem[{Balk{\i}r et~al.(2022{\natexlab{a}})Balk{\i}r, Kiritchenko,
  Nejadgholi, and Fraser}]{balkir2022trustnlp}
Esma Balk{\i}r, Svetlana Kiritchenko, Isar Nejadgholi, and Kathleen~C. Fraser.
  2022{\natexlab{a}}.
\newblock Challenges in applying explainability methods to improve the fairness
  of {NLP} models.
\newblock In \emph{Proceedings of the Second Workshop on Trustworthy Natural
  Language Processing (TrustNLP @ NAACL)}, Seattle, WA, USA.

\bibitem[{Balk{\i}r et~al.(2022{\natexlab{b}})Balk{\i}r, Nejadgholi, Fraser,
  and Kiritchenko}]{balkir-etal-2022-necessity}
Esma Balk{\i}r, Isar Nejadgholi, Kathleen Fraser, and Svetlana Kiritchenko.
  2022{\natexlab{b}}.
\newblock \href {https://aclanthology.org/2022.naacl-main.192} {Necessity and
  sufficiency for explaining text classifiers: A case study in hate speech
  detection}.
\newblock In \emph{Proceedings of the 2022 Conference of the North American
  Chapter of the Association for Computational Linguistics: Human Language
  Technologies}, pages 2672--2686, Seattle, United States. Association for
  Computational Linguistics.

\bibitem[{Belinkov(2022)}]{belinkov2022probing}
Yonatan Belinkov. 2022.
\newblock Probing classifiers: Promises, shortcomings, and advances.
\newblock \emph{Computational Linguistics}, 48(1):207--219.

\bibitem[{Beukeboom and Burgers(2019)}]{beukeboom2019stereotypes}
Camiel~J Beukeboom and Christian Burgers. 2019.
\newblock How stereotypes are shared through language: {A} review and
  introduction of the social categories and stereotypes communication {(SCSC)}
  framework.
\newblock \emph{Review of Communication Research}, 7:1--37.

\bibitem[{Blodgett et~al.(2020)Blodgett, Barocas, Daum{\'e}~III, and
  Wallach}]{blodgett2020language}
Su~Lin Blodgett, Solon Barocas, Hal Daum{\'e}~III, and Hanna Wallach. 2020.
\newblock Language (technology) is power: A critical survey of “bias” in
  {NLP}.
\newblock In \emph{Proceedings of the 58th Annual Meeting of the Association
  for Computational Linguistics}, pages 5454--5476.

\bibitem[{Borkan et~al.(2019)Borkan, Dixon, Sorensen, Thain, and
  Vasserman}]{borkan2019nuanced}
Daniel Borkan, Lucas Dixon, Jeffrey Sorensen, Nithum Thain, and Lucy Vasserman.
  2019.
\newblock Nuanced metrics for measuring unintended bias with real data for text
  classification.
\newblock In \emph{Companion Proceedings of the 2019 World Wide Web
  Conference}, pages 491--500.

\bibitem[{Castelnovo et~al.(2022)Castelnovo, Crupi, Greco, Regoli, Penco, and
  Cosentini}]{castelnovo2021zoo}
Alessandro Castelnovo, Riccardo Crupi, Greta Greco, Daniele Regoli,
  Ilaria~Giuseppina Penco, and Andrea~Claudio Cosentini. 2022.
\newblock A clarification of the nuances in the fairness metrics landscape.
\newblock \emph{Scientific Reports}, 12.

\bibitem[{Chiril et~al.(2022)Chiril, Pamungkas, Benamara, Moriceau, and
  Patti}]{chiril2022emotionally}
Patricia Chiril, Endang~Wahyu Pamungkas, Farah Benamara, V{\'e}ronique
  Moriceau, and Viviana Patti. 2022.
\newblock Emotionally informed hate speech detection: a multi-target
  perspective.
\newblock \emph{Cognitive Computation}, 14(1):322--352.

\bibitem[{Conneau et~al.(2018)Conneau, Kruszewski, Lample, Barrault, and
  Baroni}]{conneau2018you}
Alexis Conneau, German Kruszewski, Guillaume Lample, Lo{\"\i}c Barrault, and
  Marco Baroni. 2018.
\newblock What you can cram into a single vector: Probing sentence embeddings
  for linguistic properties.
\newblock \emph{arXiv preprint arXiv:1805.01070}.

\bibitem[{Czarnowska et~al.(2021)Czarnowska, Vyas, and
  Shah}]{czarnowska2021quantifying}
Paula Czarnowska, Yogarshi Vyas, and Kashif Shah. 2021.
\newblock Quantifying social biases in {NLP}: A generalization and empirical
  comparison of extrinsic fairness metrics.
\newblock \emph{Transactions of the Association for Computational Linguistics},
  9:1249--1267.

\bibitem[{Danilevsky et~al.(2020)Danilevsky, Qian, Aharonov, Katsis, Kawas, and
  Sen}]{danilevsky2020survey}
Marina Danilevsky, Kun Qian, Ranit Aharonov, Yannis Katsis, Ban Kawas, and
  Prithviraj Sen. 2020.
\newblock A survey of the state of explainable ai for natural language
  processing.
\newblock In \emph{Proceedings of the 1st Conference of the Asia-Pacific
  Chapter of the Association for Computational Linguistics and the 10th
  International Joint Conference on Natural Language Processing}, pages
  447--459.

\bibitem[{Das and Rad(2020)}]{das2020opportunities}
Arun Das and Paul Rad. 2020.
\newblock Opportunities and challenges in explainable artificial intelligence
  (xai): A survey.
\newblock \emph{arXiv preprint arXiv:2006.11371}.

\bibitem[{Dixon et~al.(2018)Dixon, Li, Sorensen, Thain, and
  Vasserman}]{dixon2018measuring}
Lucas Dixon, John Li, Jeffrey Sorensen, Nithum Thain, and Lucy Vasserman. 2018.
\newblock Measuring and mitigating unintended bias in text classification.
\newblock In \emph{Proceedings of the 2018 AAAI/ACM Conference on AI, Ethics,
  and Society}, pages 67--73.

\bibitem[{Doshi-Velez and Kim(2017)}]{doshi2017towards}
Finale Doshi-Velez and Been Kim. 2017.
\newblock Towards a rigorous science of interpretable machine learning.
\newblock \emph{arXiv preprint arXiv:1702.08608}.

\bibitem[{Elazar et~al.(2021)Elazar, Ravfogel, Jacovi, and
  Goldberg}]{elazar2021amnesic}
Yanai Elazar, Shauli Ravfogel, Alon Jacovi, and Yoav Goldberg. 2021.
\newblock Amnesic probing: Behavioral explanation with amnesic counterfactuals.
\newblock \emph{Transactions of the Association for Computational Linguistics},
  9:160--175.

\bibitem[{Fortuna and Nunes(2018)}]{fortuna2018survey}
Paula Fortuna and S{\'e}rgio Nunes. 2018.
\newblock A survey on automatic detection of hate speech in text.
\newblock \emph{ACM Computing Surveys (CSUR)}, 51(4):1--30.

\bibitem[{Ghorbani et~al.(2019)Ghorbani, Wexler, Zou, and
  Kim}]{ghorbani2019towards}
Amirata Ghorbani, James Wexler, James~Y Zou, and Been Kim. 2019.
\newblock Towards automatic concept-based explanations.
\newblock \emph{Advances in Neural Information Processing Systems}, 32.

\bibitem[{Graziani et~al.(2018)Graziani, Andrearczyk, and
  M{\"u}ller}]{graziani2018regression}
Mara Graziani, Vincent Andrearczyk, and Henning M{\"u}ller. 2018.
\newblock Regression concept vectors for bidirectional explanations in
  histopathology.
\newblock In \emph{Understanding and Interpreting Machine Learning in Medical
  Image Computing Applications}, pages 124--132. Springer.

\bibitem[{Grgic-Hlaca et~al.(2016)Grgic-Hlaca, Zafar, Gummadi, and
  Weller}]{grgic2016case}
Nina Grgic-Hlaca, Muhammad~Bilal Zafar, Krishna~P Gummadi, and Adrian Weller.
  2016.
\newblock The case for process fairness in learning: Feature selection for fair
  decision making.
\newblock In \emph{Proceedings of the NIPS Symposium on Machine Learning and
  the Law}, volume~1, page~2. Barcelona, Spain.

\bibitem[{Hanu(2020)}]{hanu_2020}
Laura Hanu. 2020.
\newblock \href
  {https://medium.com/unitary/how-well-can-we-detoxify-comments-online-bfffe5f716d7}
  {How well can we detoxify comments online?}
\newblock Unitary, accessed on 15 June, 2022.

\bibitem[{Kennedy et~al.(2020)Kennedy, Jin, Davani, Dehghani, and
  Ren}]{kennedy2020contextualizing}
Brendan Kennedy, Xisen Jin, Aida~Mostafazadeh Davani, Morteza Dehghani, and
  Xiang Ren. 2020.
\newblock Contextualizing hate speech classifiers with post-hoc explanation.
\newblock In \emph{Proceedings of the 58th Annual Meeting of the Association
  for Computational Linguistics}, pages 5435--5442.

\bibitem[{Kim et~al.(2018)Kim, Wattenberg, Gilmer, Cai, Wexler, Viegas
  et~al.}]{kim2018interpretability}
Been Kim, Martin Wattenberg, Justin Gilmer, Carrie Cai, James Wexler, Fernanda
  Viegas, et~al. 2018.
\newblock Interpretability beyond feature attribution: Quantitative testing
  with concept activation vectors (tcav).
\newblock In \emph{Proceedings of the International Conference on Machine
  Learning}, pages 2668--2677. PMLR.

\bibitem[{Kiritchenko and Mohammad(2018)}]{kiritchenko2018examining}
Svetlana Kiritchenko and Saif Mohammad. 2018.
\newblock Examining gender and race bias in two hundred sentiment analysis
  systems.
\newblock In \emph{Proceedings of the Seventh Joint Conference on Lexical and
  Computational Semantics}, pages 43--53.

\bibitem[{Koh et~al.(2020)Koh, Nguyen, Tang, Mussmann, Pierson, Kim, and
  Liang}]{koh2020concept}
Pang~Wei Koh, Thao Nguyen, Yew~Siang Tang, Stephen Mussmann, Emma Pierson, Been
  Kim, and Percy Liang. 2020.
\newblock Concept bottleneck models.
\newblock In \emph{Proceedings of the International Conference on Machine
  Learning}, pages 5338--5348. PMLR.

\bibitem[{Kumar et~al.(2022)Kumar, Tan, and Sharma}]{kumar2022probing}
Abhinav Kumar, Chenhao Tan, and Amit Sharma. 2022.
\newblock Probing classifiers are unreliable for concept removal and detection.
\newblock \emph{arXiv preprint arXiv:2207.04153}.

\bibitem[{Lundberg and Lee(2017)}]{lundberg2017unified}
Scott~M Lundberg and Su-In Lee. 2017.
\newblock A unified approach to interpreting model predictions.
\newblock \emph{Advances in Neural Information Processing Systems}, 30.

\bibitem[{Mohammad(2018)}]{vad-acl2018}
Saif~M. Mohammad. 2018.
\newblock Obtaining reliable human ratings of valence, arousal, and dominance
  for 20,000 {E}nglish words.
\newblock In \emph{Proceedings of The Annual Conference of the Association for
  Computational Linguistics (ACL)}, Melbourne, Australia.

\bibitem[{Morse et~al.(2021)Morse, Teodorescu, Awwad, and Kane}]{morse2021ends}
Lily Morse, Mike Horia~M Teodorescu, Yazeed Awwad, and Gerald~C Kane. 2021.
\newblock Do the ends justify the means? variation in the distributive and
  procedural fairness of machine learning algorithms.
\newblock \emph{Journal of Business Ethics}, pages 1--13.

\bibitem[{Nejadgholi et~al.(2022)Nejadgholi, Fraser, and
  Kiritchenko}]{nejadgholi-etal-2022-improving}
Isar Nejadgholi, Kathleen Fraser, and Svetlana Kiritchenko. 2022.
\newblock \href {https://doi.org/10.18653/v1/2022.acl-long.378} {Improving
  generalizability in implicitly abusive language detection with concept
  activation vectors}.
\newblock In \emph{Proceedings of the 60th Annual Meeting of the Association
  for Computational Linguistics (Volume 1: Long Papers)}, pages 5517--5529,
  Dublin, Ireland. Association for Computational Linguistics.

\bibitem[{Pandey(2021)}]{pandey_2021}
Parul Pandey. 2021.
\newblock \href
  {https://parulpandey.com/2019/05/09/tcav-interpretability-beyond-feature-attribution/}
  {Tcav: Interpretability beyond feature attribution}.
\newblock Breaking the Jargons, accessed on 13 June, 2022.

\bibitem[{Prabhakaran et~al.(2019)Prabhakaran, Hutchinson, and
  Mitchell}]{prabhakaran2019perturbation}
Vinodkumar Prabhakaran, Ben Hutchinson, and Margaret Mitchell. 2019.
\newblock Perturbation sensitivity analysis to detect unintended model biases.
\newblock In \emph{Proceedings of the 2019 Conference on Empirical Methods in
  Natural Language Processing and the 9th International Joint Conference on
  Natural Language Processing (EMNLP-IJCNLP)}, pages 5740--5745.

\bibitem[{Ravfogel et~al.(2020)Ravfogel, Elazar, Gonen, Twiton, and
  Goldberg}]{ravfogel2020null}
Shauli Ravfogel, Yanai Elazar, Hila Gonen, Michael Twiton, and Yoav Goldberg.
  2020.
\newblock Null it out: Guarding protected attributes by iterative nullspace
  projection.
\newblock In \emph{Proceedings of the 58th Annual Meeting of the Association
  for Computational Linguistics}, pages 7237--7256.

\bibitem[{Ribeiro et~al.(2016)Ribeiro, Singh, and Guestrin}]{ribeiro2016should}
Marco~Tulio Ribeiro, Sameer Singh, and Carlos Guestrin. 2016.
\newblock ``{W}hy should i trust you?" {E}xplaining the predictions of any
  classifier.
\newblock In \emph{Proceedings of the 22nd ACM SIGKDD International Conference
  on Knowledge Discovery and Data Mining}, pages 1135--1144.

\bibitem[{Rogers et~al.(2020)Rogers, Kovaleva, and
  Rumshisky}]{rogers2020primer}
Anna Rogers, Olga Kovaleva, and Anna Rumshisky. 2020.
\newblock A primer in bertology: What we know about how bert works.
\newblock \emph{Transactions of the Association for Computational Linguistics},
  8:842--866.

\bibitem[{R{\"o}ttger et~al.(2021)R{\"o}ttger, Vidgen, Nguyen, Waseem,
  Margetts, and Pierrehumbert}]{rottger-etal-2021-hatecheck}
Paul R{\"o}ttger, Bertie Vidgen, Dong Nguyen, Zeerak Waseem, Helen Margetts,
  and Janet Pierrehumbert. 2021.
\newblock \href {https://doi.org/10.18653/v1/2021.acl-long.4} {{H}ate{C}heck:
  Functional tests for hate speech detection models}.
\newblock In \emph{Proceedings of the 59th Annual Meeting of the Association
  for Computational Linguistics and the 11th International Joint Conference on
  Natural Language Processing (Volume 1: Long Papers)}, pages 41--58, Online.
  Association for Computational Linguistics.

\bibitem[{Selvaraju et~al.(2017)Selvaraju, Cogswell, Das, Vedantam, Parikh, and
  Batra}]{selvaraju2017grad}
Ramprasaath~R Selvaraju, Michael Cogswell, Abhishek Das, Ramakrishna Vedantam,
  Devi Parikh, and Dhruv Batra. 2017.
\newblock Grad-cam: Visual explanations from deep networks via gradient-based
  localization.
\newblock In \emph{Proceedings of the IEEE International Conference on Computer
  Vision}, pages 618--626.

\bibitem[{Shah et~al.(2020)Shah, Schwartz, and Hovy}]{shah2020predictive}
Deven~Santosh Shah, H~Andrew Schwartz, and Dirk Hovy. 2020.
\newblock Predictive biases in natural language processing models: A conceptual
  framework and overview.
\newblock In \emph{Proceedings of the 58th Annual Meeting of the Association
  for Computational Linguistics}, pages 5248--5264.

\bibitem[{Shrikumar et~al.(2017)Shrikumar, Greenside, and
  Kundaje}]{shrikumar2017learning}
Avanti Shrikumar, Peyton Greenside, and Anshul Kundaje. 2017.
\newblock Learning important features through propagating activation
  differences.
\newblock In \emph{Proceedings of the International Conference on Machine
  Learning}, pages 3145--3153.

\bibitem[{Smilkov et~al.(2017)Smilkov, Thorat, Kim, Vi{\'e}gas, and
  Wattenberg}]{smilkov2017smoothgrad}
Daniel Smilkov, Nikhil Thorat, Been Kim, Fernanda Vi{\'e}gas, and Martin
  Wattenberg. 2017.
\newblock Smoothgrad: removing noise by adding noise.
\newblock \emph{arXiv preprint arXiv:1706.03825}.

\bibitem[{Sundararajan et~al.(2017)Sundararajan, Taly, and
  Yan}]{sundararajan2017axiomatic}
Mukund Sundararajan, Ankur Taly, and Qiqi Yan. 2017.
\newblock Axiomatic attribution for deep networks.
\newblock In \emph{Proceedings of the International Conference on Machine
  Learning}, pages 3319--3328.

\bibitem[{Tenney et~al.(2019)Tenney, Xia, Chen, Wang, Poliak, McCoy, Kim,
  Van~Durme, Bowman, Das et~al.}]{tenney2019you}
Ian Tenney, Patrick Xia, Berlin Chen, Alex Wang, Adam Poliak, R~Thomas McCoy,
  Najoung Kim, Benjamin Van~Durme, Samuel~R Bowman, Dipanjan Das, et~al. 2019.
\newblock What do you learn from context? probing for sentence structure in
  contextualized word representations.
\newblock \emph{arXiv preprint arXiv:1905.06316}.

\bibitem[{Tong and Kagal(2020)}]{tong2020investigating}
Schrasing Tong and Lalana Kagal. 2020.
\newblock Investigating bias in image classification using model explanations.
\newblock In \emph{Proceedings of the ICML Workshop on Human Interpretability
  in Machine Learning (WHI 2020)}.

\bibitem[{Wei et~al.(2021)Wei, Gales, and Knill}]{wei2021analysing}
Xizi Wei, Mark~JF Gales, and Kate~M Knill. 2021.
\newblock Analysing bias in spoken language assessment using concept activation
  vectors.
\newblock In \emph{Proceedings of the IEEE International Conference on
  Acoustics, Speech and Signal Processing (ICASSP)}, pages 7753--7757. IEEE.

\bibitem[{Yeh et~al.(2020)Yeh, Kim, Arik, Li, Pfister, and
  Ravikumar}]{yeh2020completeness}
Chih-Kuan Yeh, Been Kim, Sercan Arik, Chun-Liang Li, Tomas Pfister, and Pradeep
  Ravikumar. 2020.
\newblock On completeness-aware concept-based explanations in deep neural
  networks.
\newblock \emph{Advances in Neural Information Processing Systems},
  33:20554--20565.

\bibitem[{Yeh et~al.(2022)Yeh, Kim, and Ravikumar}]{hitzler2022human}
Chih-Kuan Yeh, Been Kim, and Pradeep Ravikumar. 2022.
\newblock Human-centered concept explanations for neural networks.
\newblock In P.~Hitzler and M.~K. Sarker, editors, \emph{Neuro-Symbolic
  Artificial Intelligence: The State of the Art}, volume 342, page~2. IOS
  Press.

\bibitem[{Zhou et~al.(2021)Zhou, Yong, Fan, Ren, Song, Diao, Yang, and
  Lin}]{zhou-etal-2021-hate}
Xianbing Zhou, Yang Yong, Xiaochao Fan, Ge~Ren, Yunfeng Song, Yufeng Diao,
  Liang Yang, and Hongfei Lin. 2021.
\newblock \href {https://doi.org/10.18653/v1/2021.acl-long.556} {Hate speech
  detection based on sentiment knowledge sharing}.
\newblock In \emph{Proceedings of the 59th Annual Meeting of the Association
  for Computational Linguistics and the 11th International Joint Conference on
  Natural Language Processing (Volume 1: Long Papers)}, pages 7158--7166,
  Online. Association for Computational Linguistics.

\end{thebibliography}
\bibliographystyle{acl_natbib}

\appendix

\section{Sensitivities to Sentiment in Presence of Identity Terms}
\label{sec:appendix}

The full results of the experiments described in Section~\ref{sec:group-sentiment} are presented in Table~\ref{tab:sentiment_subjects}. 

\setcounter{table}{0}
\renewcommand\thetable{A.\arabic{table}}

\begin{table*}[!ht]
    \centering
    \small{
    \begin{tabular}{c|c|c|c|c|c|c}
    \hline
  Target&Class label &Very negative&Negative&Neutral&Positive& Very positive\\
         
     \hline

  \hline
  \hline

      \multirow{7}{*}{Women} & \textit{Toxicity}&\textbf{0.99(0.00)}& \textbf{0.99(0.00)}& 0.24(0.22)& 0(0)& 0(0) \\
       \cline{2-7}
       &\textit{Obscene}   & 0(0)& 0(0)&0(0)& 0(0)& 0(0)\\
       \cline{2-7}
       &\textit{Identity Attack} & \textbf{1.00(0)}&\textbf{0.99(0.00)}&\textbf{0.98(0.02)}& 0(0.01)& 0(0)\\
     \cline{2-7}
       & \textit{Insult}& \textbf{0.98(0.01)}& \textbf{0.87(0.11)}&0(0)&0(0)&0(0)\\
       \cline{2-7}
       & \textit{Threat} &0(0)& 0(0)&0(0)& 0(0)& 0(0)\\
       \cline{2-7}
       & \textit{Sexual Explicit} &  0(0)& 0(0)&0(0)& 0(0)& 0(0)\\
       
        \hline

  \hline
  \hline
      \multirow{7}{*}{Trans people} & \textit{Toxicity} &\textbf{0.97(0.01)}&\textbf{0.93(0.01)}&\textbf{0.78(0.04)}&0(0)&0(0)\\
      \cline{2-7}
      & \textit{Obscene}   & 0(0)&0(0)&0(0)&0(0)
&0(0)\\
      \cline{2-7}
      &\textit{Identity Attack} & \textbf{1.00(0)}&\textbf{1.00(0)}&\textbf{0.92(0.01)}&\textbf{0.32(0.20)}&0.02(0.05)\\
    \cline{2-7}
      & \textit{Insult}& \textbf{0.97(0.007)}&\textbf{0.94(0.01)}&\textbf{0.77(0.07)}&0(0)&0(0)\\
      \cline{2-7}
      & \textit{Threat}&0(0)&0(0)&0(0)&0(0)&0(0)\\
      \cline{2-7}
        &\textit{Sexual Explicit} &0(0)&0(0)&0(0)&0(0)&0(0)\\
     
     \hline

  \hline
  \hline
       \multirow{7}{*}{Gay people} & \textit{Toxicity} &\textbf{1.00(0)}&\textbf{0.99(0.00)}&\textbf{0.99(0.00)}&\textbf{0.97(0.00)}&\textbf{0.93(0.01)}\\
       \cline{2-7}
      & \textit{Obscene}   & 0.25(0.12)&0.01(0.02)&0(0)&0(0)
&0(0)\\
       \cline{2-7}
       &\textit{Identity Attack} & \textbf{1.00(0)}&\textbf{1.00(0)}&\textbf{0.99(0.00)}&\textbf{0.99(0.00)}&\textbf{0.99(0.00)}\\
    \cline{2-7}
       & \textit{Insult}& \textbf{0.99(0.001)}&\textbf{0.99(0.003)}&\textbf{0.91(0.04)}&0(0)&0(0)\\
       \cline{2-7}
       & \textit{Threat}&0(0)&0(0)&0(0)&0(0)&0(0)\\
       \cline{2-7}
        &\textit{Sexual Explicit} & \textbf{0.82(0.02)}&\textbf{0.84(0.01)}&\textbf{0.88(0.01)}&
\textbf{0.82(0.02)}&\textbf{0.73(0.02)}\\
     
     \hline

  \hline
  \hline
      
        \multirow{7}{*}{Black people} & \textit{Toxicity} &\textbf{1.00(0)}&\textbf{1.00(0)}&\textbf{0.99(0.00)}&\textbf{0.95(0.00)}&\textbf{0.92(0.01)}\\
      \cline{2-7}
      & \textit{Obscene}   & 0.05(0.06)&0.00(0.00)&0(0)&0(0)
&0(0)\\
      \cline{2-7}
      &\textit{Identity Attack} & \textbf{1.00(0)}&\textbf{1.00(0)}&\textbf{1.00(0)}&\textbf{0.99(0.00)}&\textbf{0.99(0.00)}\\
    \cline{2-7}
      & \textit{Insult}& \textbf{0.99(0.001)}&\textbf{0.99(0.002)}&\textbf{0.95(0.01)}&0.14(0.12)&0(0)\\
      \cline{2-7}
      & \textit{Threat}&0.03(0.02)&0.04(0.02)&0(0)&0(0)&0(0)\\
      \cline{2-7}
        &\textit{Sexual Explicit} &0(0)&0(0)&0(0)&0(0)&0(0)\\
     
     \hline

  \hline
  \hline
        \multirow{7}{*}{Disabled people} & \textit{Toxicity} &\textbf{0.41(0.2)}&0.01(0.06)&0(0)&0(0)&0(0)\\
      \cline{2-7}
      & \textit{Obscene}    &0(0)&0(0)&0(0)&0(0)&0(0)\\
      \cline{2-7}
      &\textit{Identity Attack} &0(0)&0(0)&0(0)&0(0)&0(0)\\
    \cline{2-7}
      & \textit{Insult}& \textbf{0.70(0.2)}&0.13(0.21)&0(0)&0(0)&0(0)\\
      \cline{2-7}
      & \textit{Threat}&0(0)&0(0)&0(0)&0(0)&0(0)\\
      \cline{2-7}
        &\textit{Sexual Explicit} &0(0)&0(0)&0(0)&0(0)&0(0)\\
     
     \hline

  \hline
  \hline
    \multirow{7}{*}{Muslims} & \textit{Toxicity} &\textbf{1.00(0)}&\textbf{0.99(0.00)}&\textbf{0.96(0.01)}&\textbf{0.75(0.04)}&\textbf{0.57(0.07)}\\
      \cline{2-7}
      &\textit{Obscene}   & 0(0.02)&0(0)&0(0)&0(0)
&0(0)\\
      \cline{2-7}
      &\textit{Identity Attack} & \textbf{1.00(0)}&\textbf{1.00(0)}&\textbf{0.99(0.00)}&\textbf{0.98(0.00)}&\textbf{0.97(0.00)}\\
     \cline{2-7}
      &  \textit{Insult}& \textbf{0.99(0.00)}&\textbf{0.98(0.007)}&\textbf{0.69(0.15)}&0(0)&0(0)\\
       \cline{2-7}
        &\textit{Threat}&\textbf{0.33(0.07)}&\textbf{0.32(0.07)}&\textbf{0.20(0.06)}&0(0)&0(0)\\
        \cline{2-7}
        &\textit{Sexual Explicit} & 0(0)&0(0)&0(0)&
0(0)&0(0)\\
     \hline
     
  \hline
  \hline
        \multirow{7}{*}{Immigrants} & \textit{Toxicity} &\textbf{0.98(0)}&\textbf{0.95(0.01)}&\textbf{0.86(0.03)}&0(0)&0(0)\\
      \cline{2-7}
      & \textit{Obscene}  & 0(0)&0(0)&0(0)&0(0)&0(0)\\
      \cline{2-7}
      &\textit{Identity Attack} & \textbf{1.00(0)}&\textbf{0.99(0)}&\textbf{0.95(0.01)}&\textbf{0.74(0.11)}&\textbf{0.30(0.23)}\\
    \cline{2-7}
      & \textit{Insult}& \textbf{0.98(0.005)}&\textbf{0.96(0.01)}&\textbf{0.86(0.03)}&0(0)&0(0)\\
      \cline{2-7}
      & \textit{Threat}&0(0)&0(0)&0(0)&0(0)&0(0)\\
      \cline{2-7}
        &\textit{Sexual Explicit} & 0(0)&0(0)&0(0)&0(0)&0(0)\\
     
     \hline
     
  \hline
  \hline
        \multirow{7}{*}{These people} & \textit{Toxicity} &\textbf{0.93(0.02)}&\textbf{0.83(0.07)}&\textbf{0(0.03)}&0(0)&0(0)\\
      \cline{2-7}
      & \textit{Obscene}  & 0(0)&0(0)&0(0)&0(0)&0(0)\\
      \cline{2-7}
      &\textit{Identity Attack}& 0(0)&0(0)&0(0)&0(0)&0(0)\\
    \cline{2-7}
      & \textit{Insult}& \textbf{0.97(0.01)}&\textbf{0.92(0.02)}&0.13(0.21)&0(0)&0(0)\\
      \cline{2-7}
      & \textit{Threat}&0(0)&0(0)&0(0)&0(0)&0(0)\\
      \cline{2-7}
        &\textit{Sexual Explicit} & 0(0)&0(0)&0(0)&0(0)&0(0)\\
     
     \hline

  \hline
  \hline
        \multirow{7}{*}{These things} & \textit{Toxicity}& 0(0)&0(0)&0(0)&0(0)&0(0)\\
     
      \cline{2-7}
     
      & \textit{Obscene}  & 0(0)&0(0)&0(0)&0(0)&0(0)\\
     
      \cline{2-7}
      &\textit{Identity Attack}& 0(0)&0(0)&0(0)&0(0)&0(0)\\
     
    \cline{2-7}
      & \textit{Insult}& 0(0.03)&0(0)&0(0)&0(0)&0(0)\\
     
      \cline{2-7}
      & \textit{Threat}&0(0)&0(0)&0(0)&0(0)&0(0)\\
      \cline{2-7}
        &\textit{Sexual Explicit} & 0(0)&0(0)&0(0)&0(0)&0(0)\\
     
     \hline
    \end{tabular}
    \caption{Average and standard deviation of TCAV scores for all the labels and different levels of sentiment ranging from Very Negative to Very Positive for the template  \textbf{``{\small<SUBJECTS>} are {\small<SENTIMENT-WORD>}''}. All the sensitivities that are significantly different from random are in bold.  }
    \label{tab:sentiment_subjects}
    }
\end{table*}

\end{document}